# Establishing Causal Relationship Between Whole Slide Image Predictions and Diagnostic Evidence Subregions in Deep Learning

Tianhang Nan[1], Yong Ding[1], Hao Quan[1], Deliang Li[1], Lisha Li[1], Guanghong Zhao[1] and Xiaoyu Cui[*1, 2]

[1]College of Medicine and Biological Information Engineering, Northeastern University, China
[2]The Key Laboratory of Biomedical Imaging Science and System, Chinese Academy of Sciences, China

e-mail: tianhang_nan@foxmail.com (Tianhang Nan), dingyong.prc@foxmail.com (Yong Ding), h.quan@siat.ac.cn (Hao Quan), fengya123xyn@foxmail.com (Deliang Li), 18040105616@163.com (Lisha Li); 18854803079@163.com (Guanghong Zhao); cuixy@bmie.neu.edu.cn (Xiaoyu Cui, the corresponding author).

***Abstract*****—**Due to the lack of fine-grained annotation guidance, current Multiple Instance Learning (MIL) struggles to establish a robust causal relationship between Whole Slide Image (WSI) diagnosis and evidence sub-images, just like fully supervised learning. So many noisy images can undermine the network's prediction. The proposed Causal Inference Multiple Instance Learning (CI-MIL), uses out-of-distribution generalization to reduce the recognition confusion of sub-images by MIL network, without requiring pixelwise annotations. Specifically, feature distillation is introduced to roughly identify the feature representation of lesion patches. Then, in the random Fourier feature space, these features are re-weighted to minimize the cross-correlation, effectively correcting the feature distribution deviation. These processes reduce the uncertainty when tracing the prediction results back to patches. Predicted diagnoses are more direct and reliable because the causal relationship between them and diagnostic evidence images is more clearly recognized by the network. Experimental results demonstrate that CI-MIL outperforms state-of-the-art methods, achieving 92.25% accuracy and 95.28% AUC on the Camelyon16 dataset (breast cancer); while 94.29% accuracy and 98.07% AUC on the TCGA-NSCLC dataset (non-small cell lung cancer). Additionally, CI-MIL exhibits superior interpretability, as its selected regions demonstrate high consistency with ground truth annotations, promising more reliable diagnostic assistance for pathologists.

## 1. INTRODUCTION

PATHOLOGY serves as the "gold standard" for preoperative diagnosis, with its diagnostic outcomes accurately unveiling the disease category and progression of patients [1-3], signifying a significant contribution to patient care. The exploration to enhance the accuracy of pathological diagnosis has therefore provided immense assistance in healthcare management. In recent years, advancements in digital pathology and artificial intelligence have enabled histopathological slides to be scanned and digitized into gigapixel Whole Slide Images (WSI) [4, 5], thrusting the domain into a computational analysis era. The emergence of this interdisciplinary field has transformed the traditional paradigm of pathology diagnosis conducted solely by pathologists under a microscope, addressing challenges related to scarcity of pathology expertise, prolonged diagnostic durations, and insufficient diagnostic accuracy [6-8]. Beyond direct clinical benefits, computational pathology has shown potential in tasks such as microenvironment analysis [9, 10] and drug testing [11].

Recently, weakly supervised learning [1, 12] methods represented by Multiple Instance Learning (MIL) [13] have received widespread attention in the field of pathological image classification. These methods only use reported pathological diagnoses as labels for training without the need for manual pixelwise annotation [14-17]. Compared with fully supervised learning, MIL can save a large amount of medical human resources. MIL-based methods partition WSIs into a vast number of patches, using slide-level labels as "pseudo-labels" for the patches, enabling the model to diagnose or extract features from patches and ultimately aggregate the results of all patches to obtain WSI predictions [18-23]. Guided by this principle, these methods primarily focus on better feature fusion mechanisms.

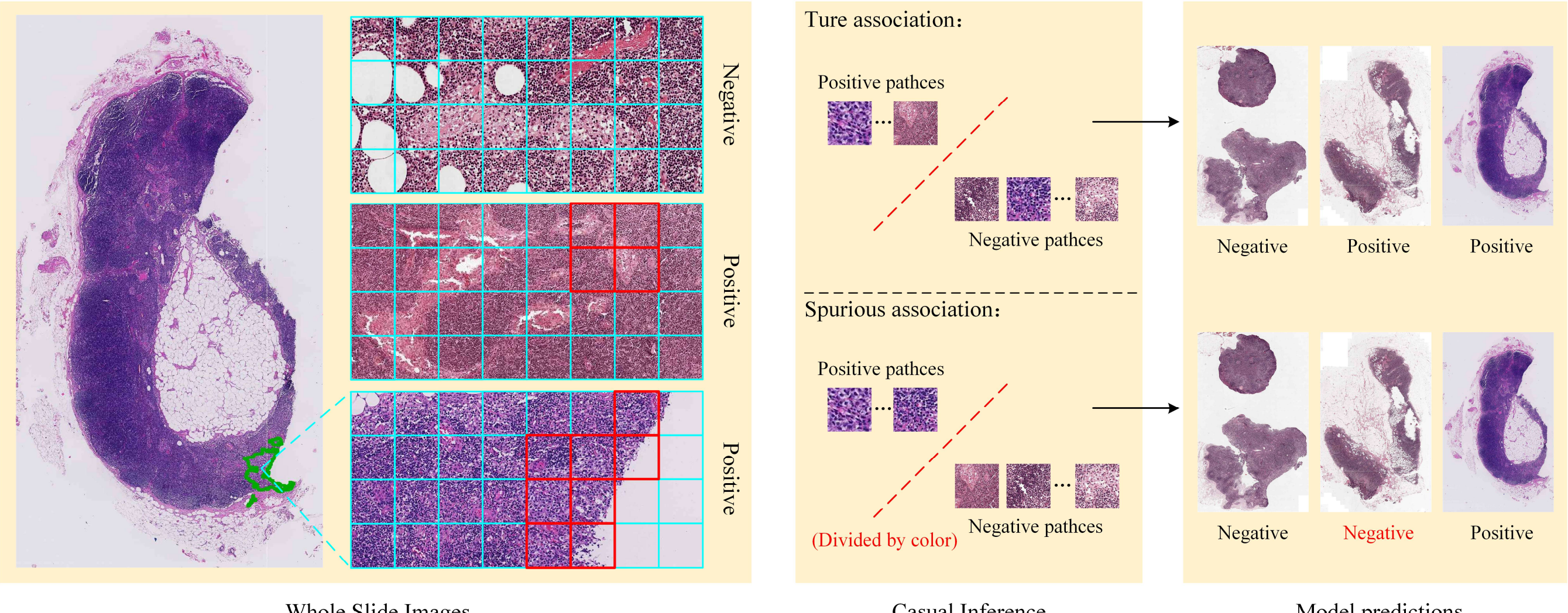

Fig. 1. Spurious causal relationships existing in WSI classification models.

However, these MIL-based research neglects some important issues. As shown in Fig.1, slide-level labels correspond to only minute regions within each gigapixel image [2]. At the same time, due to the complexity of pathological images and lack of fine-grained annotation guidance, patch-level diagnostic results often correlate with incorrect image features. For example, color errors caused by uneven staining, different scanning equipment, can lead to models introducing incorrect prediction bases [22, 24], while pathologists usually only consider cell morphology. As shown in Fig.2.a, after segmenting WSIs into patches, the features obtained through image encoding are distributed on a hyperplane. The task for deep learning models is to find an appropriate separation on this hyperplane (taking a binary classifier as an example). We assume that this hyperplane can be mapped to a three-dimensional space understandable to human cognition, where the xy-plane represents the feature distribution space of patches, and the z-axis represents the conditional probability of these features belonging to positive or negative instances. A robust feature hyperplane should be smooth and steady, without the presence of "waves". The reason is that when the distributions (positions on the xy-plane) of two features are close, their corresponding probabilities should also be close. When there is false correlation between patch-level features, they can get similar z-values even with large differences in their feature distributions. In this case, prediction errors may occur at the "peaks" and "valleys". Constrained by performance, the distributions learned by existing deep learning models will inevitably differ from the real-world. A single WSI contains thousands of patches, and most of them are not related to diagnosis. The noise they generate is tremendous. Typically, research based on improved aggregators, as depicted in Fig.2.b, attempts to fit more complex separation curves to handle the feature hyperplane [1, 2, 18-20]. However, such methods suffer from three common issues: 1. Subregions of the hyperplane belonging to the same class might be multiple and non-contiguous, while classifiers usually delineate single contiguous regions. 2. Overfitting may occur in the vicinity of complex separation curves. 3. The inability to address the problem of false correlations between diagnostic results and specific patches (e.g., caused by color variations).

An important method for addressing these issues involves achieving out-of-distribution generalization by decorrelating the features of diagnostic evidence patches and non-diagnostic evidence patches. Due to the absence of additional supervision (pixelwise annotation) for segregating the feature representations of these two types of patches, a conservative and effective approach is to decorrelate all features. This concept has been proven to be effective in improving the model's prediction ability [25-27] and can introduce the causal inference framework into the field of WSI classification. That is, a classifier that can clearly distinguish the differences in most patch features can effectively locate the cause of model prediction in diagnostic evidence patches, thereby achieving better performance and interpretability. Explained from the huge and redundant data characteristics of WSI, the causal inference framework is expected to reduce the noise impact of non-lesion areas. As shown in Figure 2.c, we first delete the easily classified flat areas in the hyperplane through multi-channel feature extraction. Feature distillation is defined as extracting patches of interest in the WSI classification task, to eliminate interferences in WSI preliminarily [21]. Due to the nature of neural networks, the feature distributions of the output patches are similar and difficult to distinguish. Feature distillation effectively constrains the subsequent network to focus on handling this difficult problem. For example, for a WSI diagnosed as positive by a pathologist, a subset of patches that the model considers most likely to be positive in this WSI is selected for further processing. In this way, the flat sub-regions belonging to the negative class within the hyperplane are excluded. By excluding patches that are clearly irrelevant to the prediction, feature distillation establishes an initial connection between model prediction and evidence. Subsequently, the distilled features are mapped to random Fourier space to obtain learnable weights that minimize the interdependence of patch features (this mapping can address complex nonlinear correlations). By reweighting the features, a smoother hyperplane with reduced interdependence is obtained. Lower feature interdependence reduces situations where features with different distributions on the original hyperplane have the same conditional probability. The model can more easily recognize patches that are very close but represent different diagnostic categories. The multi-channel hyperplanes are amalgamated into a single hyperplane. On such a smooth hyperplane, a classification curve that is more akin to the real-world, simpler, and avoids overfitting and false correlations can be fitted. Finally, the superior patch-level features obtained are fused to construct a WSI classifier. The method proposed herein is termed Causal Inference Multiple Instance Learning (CI-MIL).

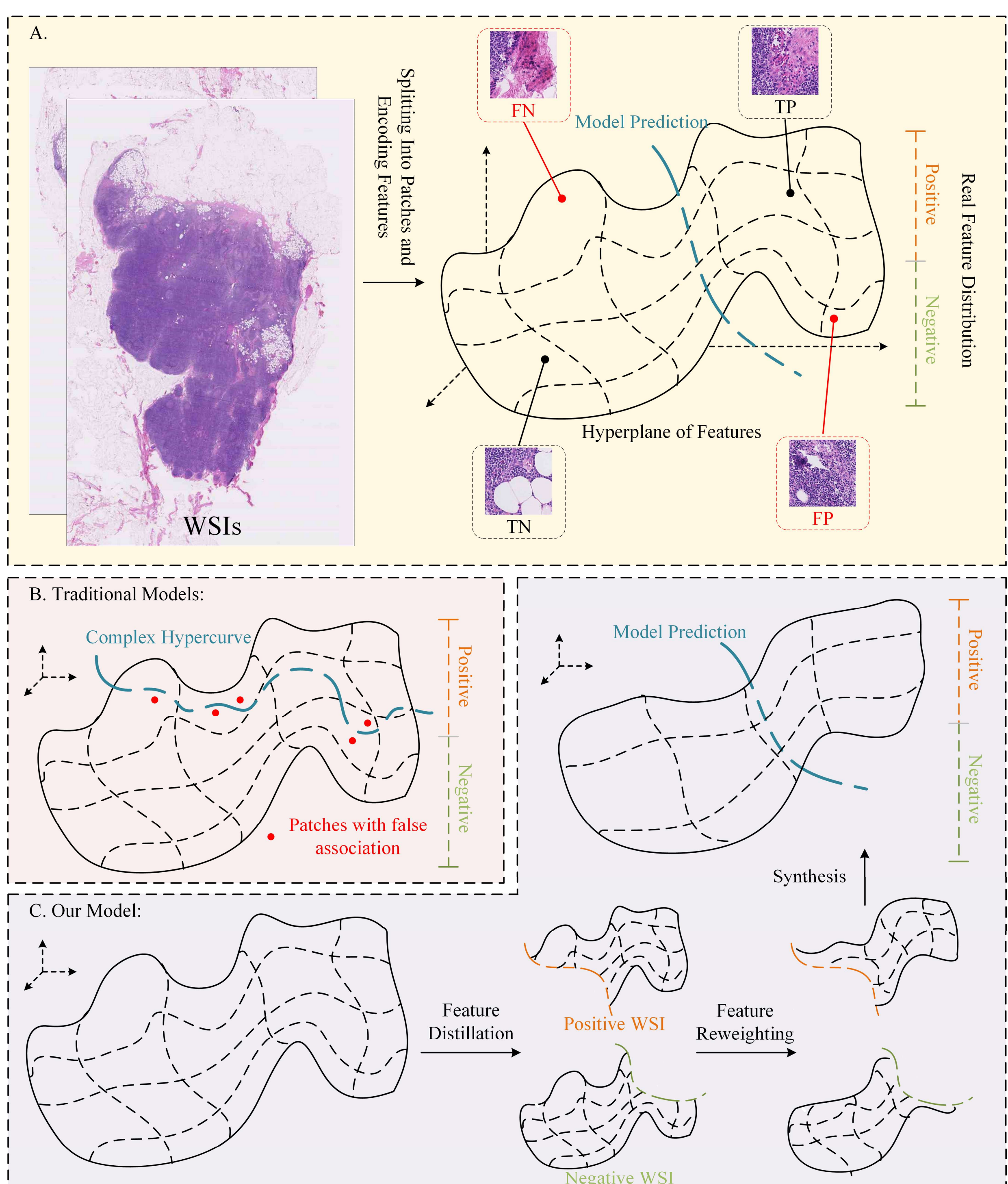

Fig.2. a. The patch-level feature hyperplane in WSI classification tasks has a deviation from the real world; b. Mainstream methods usually improve accuracy through more complex partitioning; c. This study proposes to reshape the feature hyperplane to make it closer to the real-world distribution, thereby improving classification performance.

CI-MIL, through its two-stage, multi-channel method of causal inference, effectively harnesses patches with genuine diagnostic relevance to achieve diagnostic predictions for WSIs. The approach to addressing the problem places emphasis on feature decorrelation and relationship between diagnostic evidence subregions (cause) and predictions (effect), distinguishing it from prior weakly supervised learning studies in the field, resulting in higher performance and interpretability. It's worth noting that a substantial amount of research has already explored advanced feature fusion methods. However, our research focus is entirely decoupled from this aspect. CI-MIL is a universal model that can be organically integrated with existing models to consistently enhance performance. We rigorously evaluated CI-MIL on the Camelyon16 (Camelyon Challenge 2016, breast cancer) [28] and TCGA-NSCLC (The Cancer Genome Atlas Program: non-small cell lung cancer) [29] datasets under stringent, non-stationary external conditions, surpassing existing state-of-the-art models. On the Camelyon16 dataset, CI-MIL achieved an ACC of 92.25% and an AUC of 95.28%. Additionally, on the TCGA-NSCLC dataset, it achieved 94.29% ACC and 98.07% AUC. Furthermore, CI-MIL autonomously identifies regions with high diagnostic relevance, demonstrating a high degree of consistency with the ground truth. The code for this study was made public at https://github.com/MasyerN/CI-MIL. The primary contributions of this research are as follows:

1) We propose the issue of distribution bias and false correlation in the patch-level feature hyperplane in WSI classification tasks based on weakly supervised learning, and analyze the optimization flaws of existing studies towards this hyperplane.
2) Feature distillation and removal of non-linear feature correlation are introduced to reorganize the patch-level feature hyperplane. The distribution of features is optimized to construct a causal relationship between diagnostic evidence sub-region images and model predictions.
3) Under the robust causal relationship between prediction and evidence, CI-MIL demonstrates high interpretability. While, it is orthogonal to existing studies, thereby obtaining rich extensions and achieving consistently high performance in various environments.

## 2. Related Work

Variants of Multi-Instance Learning (MIL) based on weakly supervised learning have completely transformed the WSI auxiliary diagnosis task [28-32], as they do not require expensive manual pixel-by-pixel annotation [34-36]. MIL methods define a WSI as a "bag" containing multiple instances. By assigning "pseudo-labels" to instances, extracting their features or diagnostic predictions and fusing them, a bag-level prediction result can be obtained. Guided by this principle, these methods mainly focus on better feature fusion mechanisms. Common practices include: multi-scale image aggregators [18]; trainable attention weights [19]; and the use of transformer-based self-attention mechanisms [20], which are currently the most advanced feature fusion mechanisms.

Causal inference [25-27] has received widespread attention in recent years, with a key advantage being the ability to eliminate the harmful confounding effects of irrelevant features on model predictions through additional interventions [37, 38]. Given the vast and complex features of pathological images, causal inference can reduce the interference caused by overly redundant patches and has been preliminarily introduced into this field. Some studies [22,23] have attempted to achieve external generalization of WSI data through feature clustering. Under the concept of the patch-level feature hyperplane discussed in Chapter 1, these methods modify the classification curve rather than the feature hyperplane, and cannot eliminate the existing bias in the real-world distribution through clustering. Therefore, CI-MIL designs a complex causal inference process centered around eliminating feature inter-correlation, which can better establish a real association between model predictions and diagnostic evidence subregions.

## 3. Method

The construction of CI-MIL will be discussed in four main sections: 1. Task setting and image preprocessing; 2. Feature distillation based on patch-level classification; 3. Patch-feature decorrelation in random Fourier space; 4. Feature fusion and WSI-level classification. As shown in Fig.3.a, first, we segment the WSI into organ regions to remove the background, and divide the foreground into several patches. Then, we use a pre-trained convolutional neural network to encode the images into features. Based on weakly supervised learning, we construct a patch-level classifier for feature distillation (Fig.3.b), selecting unstable patch features for further processing. These features are mapped into random Fourier space, where weights that minimize their intercorrelation are learned via gradient descent and are used for updates (Fig.3.c). The features in this new space are fused to obtain WSI-level features, and finally, a WSI classifier is constructed.

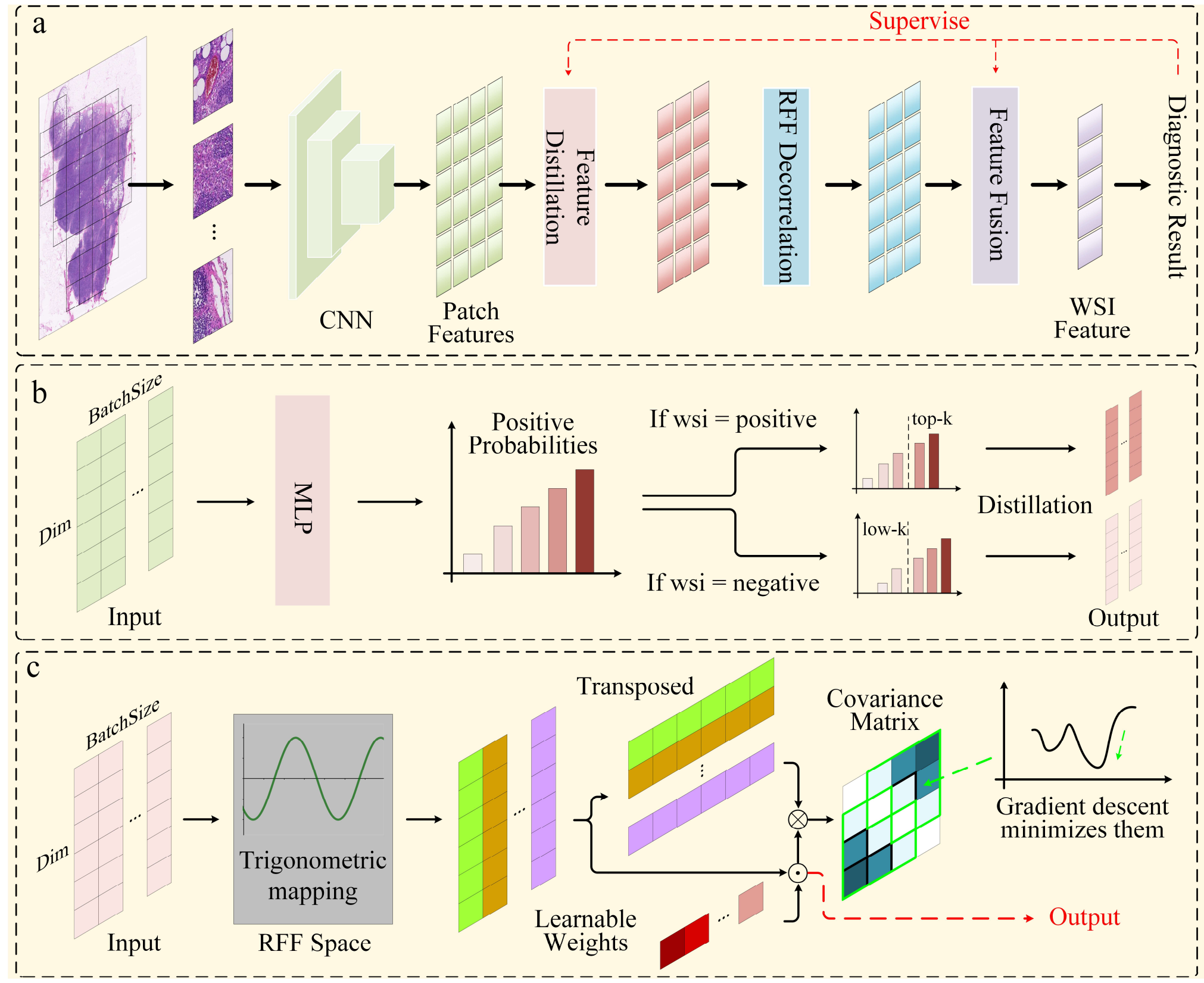

Fig. 3. a. Overall structure of CI-MIL; b. Feature distillation module; c. Patch-level feature decorrelation in random Fourier feature space.

## 3.1 Task Setting and Image Preprocessing

For a given WSI $W$, we segment it into foreground and background parts using an Otsu-based tissue region segmentation algorithm consistent with previous research. Subsequently, the foreground is divided into patches of 256×256-pixel size to fit the input size of the neural network. Consistent with previous research, a pre-trained image encoder is used to extract features from the patches. The details are described in Section 4.1.

The patches yield instance features $X = \{x_1, x_2, \ldots, x_K\}$, where $K$ represents the total number of patches contained in $W$. Each individual patch $x_i$ has a latent label $y_i$ (where $y_i = 1$ denotes positive, and $y_i = 0$ denotes negative). These labels exist in reality but are unknown to the model. The labels Y for WSI W are known to the model and are represented as:

$$Y = \begin{cases} 1, if \sum_{i=1}^{K} y_i > 0 \\ 0, if \sum_{i=1}^{K} y_i = 0 \end{cases} \tag{1}$$

When at least one patch in WSI is classified as positive, the WSI is categorized as positive; otherwise, the WSI is classified as negative.

## 3.2 Feature distillation

Feature distillation is an important initial step for constructing the correlation between model prediction and diagnostic evidence. From the perspective of model construction, the nature of neural networks makes it output patches with similar feature distributions, which can provide a relatively stable environment for subsequent patch-feature decorrelation. Performing decorrelation on all patch features without preprocessing will cause the model to ignore those similar patches that are difficult to be separated. Because generalizing patches with significantly different feature distributions is relatively simple, it easily leads to "shortcut learning" of the model [39]. From the perspective of the huge and complex data characteristics of WSI, feature distillation can exclude redundant patches unrelated to diagnosis and construct a preliminary causal relationship.

Feature distillation is implemented by a patch-level classification network, which is denoted as $f_{distillation}$. Its task is to separately select several instances from $W$ that have the highest probability of being positive and negative, respectively. As shown in Fig.3.b, $f_{distillation}$ will predict whether each instance $x_i$ belongs to the tumor:

$$\hat{y}_i = f_{distillation}(x_i) \tag{2}$$

Where $\hat{y}_i$ represents the probability of $x_i$ being predicted as positive. Once predictions for all patches in $W$ are made, the labels $Y$ are used to calculate the loss by considering the top-$k$ highest predicted probabilities $\{\hat{h}_1, \hat{h}_2, \dots \hat{h}_k\}$ from $\{\hat{y}_1, \hat{y}_2, \dots, \hat{y}_K\}$. The gradient is then backpropagated accordingly, enabling the weakly supervised learning of $f_{distillation}$:

$$\hat{h}_1, \hat{h}_2, \dots \hat{h}_k = \text{argmax}(\hat{y}_1, \hat{y}_2, \dots \hat{y}_K, k) \tag{3}$$

$$loss_1 = -\frac{1}{k}\sum_{i=0}^{k} Y * \log(\hat{h}_i) + (1 - Y) * \log(1 - \hat{h}_i) \tag{4}$$

$$\{h_1, h_2, \dots, h_k\} = \arg\left(\hat{h}_1, \hat{h}_2, \dots \hat{h}_k\right), h \epsilon \mathbb{R}^{n \times 1} \tag{5}$$

where $\{h_1, h_2, \dots, h_k\}$ represents the patch features distilled by $f_{distillation}$.

For benign-malignant classification tasks (such as Camelyon16), positive patches can represent definitive diagnostic evidence. However, for tumor subtype classification tasks, representative patches belonging to both negative and positive need to be considered (for example, in TCGA-NSCLC, negative and positive represent different tumor subtypes, and it is not enough to simply distill representative positive patches). Therefore, we simultaneously select the top $k/2$ features with the highest positive probability and the top $k/2$ features with the highest negative probability. These features are jointly used for subsequent feature fusion and WSI classification tasks:

$$\begin{aligned}\{\hat{h}_1, \hat{h}_2, \dots \hat{h}_k\} = \text{argmax}\left(\hat{y}_1, \hat{y}_2, \dots \hat{y}_K, \frac{k}{2}\right), \\ \text{argmin}\left(\hat{y}_1, \hat{y}_2, \dots \hat{y}_K, \frac{k}{2}\right)\end{aligned} \tag{6}$$

$$\{h_1, h_2, \dots, h_k\} = \arg\left(\hat{h}_1, \hat{h}_2, \dots \hat{h}_k\right) \tag{7}$$

### 3.3 Patch Feature Decorrelation in Random Fourier Feature Space

Patch feature decorrelation is to enable the model to more clearly identify the differences between patches. The tiny differences in the feature space make these patches easily recognized as "approximate" by neural networks. In reality, they may have different diagnoses. As shown in the feature hyperplane in Fig 2.a, the prediction of WSI may be associated with patches that are predicted incorrectly, which has a serious risk in medical tasks. This challenge is difficult to handle only by improving the feature fusion method [22, 38].

As shown in Fig.3.c, decorrelation between different patch features is achieved by globally weighting patch features. Specifically, the two-dimensional features output by the previous layer of the neural network are mapped to a three-dimensional space through Fourier transformation and multiple random sampling. Subsequently, using gradient descent and learnable weights, the sum of the off-diagonal elements of the feature correlation matrix is minimized, and the processed features are output. In order to expand the decorrelation operation within a batch to all samples, to adapt to the hardware limitations of training deep learning models, we further introduce a mechanism to save and reload global correlation calculations.

#### 3.3.1 Random Fourier Feature Space

For the distilled features $\{h_1, \dots\}$, the mapping $H_{RFF}$ to random Fourier Feature (RFF) space is given by the following formula:

$$\begin{aligned}H_{RFF} = \{o: h \rightarrow \sqrt{2}cos(\omega h + \phi) \mid \omega \sim N(0, 1), \\ \phi \sim \text{Uniform}(0, 2\pi)\}\end{aligned} \tag{8}$$

$$\{o_1, o_2, \dots, o_k\} = H_{RFF}(h_1, h_2, \dots, h_k), o \epsilon \mathbb{R}^{n \times m} \tag{9}$$

Where $\omega$ is sampled from a standard normal distribution, and $\phi$ is sampled from a uniform distribution. After this mapping, we obtain a high-dimensional feature space $\{o_1, o_2, \dots, o_k\}$with dimensions$k \times n \times m$, where $m$ is the number of random samplings. For the sake of brevity, we set it to 1 in the subsequent derivations.

3.3.2 Weight Optimization

Subsequent derivations should be regarded as having weighted the patches representing different diagnostic categories respectively to ensure that the feature distributions within the batch are close. This setup will not be repeatedly described, for the sake of brevity.

Each patch feature $o_i$ is assigned a learnable weight $u_i$, and is thus reweighted:

$$o_i' = u_i * o_i, \sum_{i=1}^{k} u_i \equiv k \tag{10}$$

Subsequently, we calculate the weighted correlation matrix, which is a real matrix of size $k \times k$. Each element in the matrix represents the intercorrelation of features within the batch. The closer the value of an element is to 0, the lower the correlation.

We adopt two methods to calculate the correlation matrix. Inspired by statistics, calculating the covariance matrix $[Cov]$ can well represent the correlation between features. Any element in $[Cov]$ is calculated as follows:

$$[Cov]_{i,j} = \frac{1}{n-1} \sum_{p=1}^{n} \left(o_{i,p}' - \overline{o_i'}\right) * \left(o_{j,p}' - \overline{o_j'}\right) \tag{11}$$

where $\overline{o_i'}$ represents the mean value of the elements of the corresponding feature $o_i'$.

Additionally, when the feature vectors of two patches are orthogonal in space, they also exhibit non-correlation. Therefore, by constructing the correlation matrix through vector inner products and optimizing the weights $\{u_1, u_2, \dots, u_k\}$, so that the inner products between different features approach zero, we can achieve the goal of reducing correlation. The correlation matrix $[In - product]$ constructed in this way can be represented as:

$$[In - product] = \{o_1, o_2, \dots, o_k\}^{\top} \times \{o_1', o_2', \dots, o_k'\} \tag{12}$$

By using the gradient descent algorithm, we can make the sum of the off-diagonal elements of the correlation matrix as small as possible, thereby significantly reducing the dependence between features:

$$loss_2 = \sum_{1 \le i,j \le k} [Cov]_{i,j} \ or \sum_{1 \le i,j \le k} [In - product]_{i,j} \, ; i \neq j \tag{13}$$

$$\{u_1, u_2, \dots, u_k\}^* = \operatorname{argmin}(loss_2), \sum_{i=1}^{k} u_i \equiv k \tag{14}$$

3.3.3 Global Weight Optimization

In the context of deep learning tasks, decorrelating all samples simultaneously requires substantial computational resources. And only a portion of the samples can be observed in each batch, which obviously has a bias compared to the real-world feature distribution. Therefore, we adopted a method to save and reload decorrelation calculations. This method can merge and save all features and learnable weights encountered during all training stages, and reload them before starting the decorrelation calculation for a new batch. This method has been proven to effectively extend the patch-level feature decorrelation operation within a batch to all training data, enabling the model to learn global knowledge [38].

Before the decorrelation operation for each batch $\{O = [o_1, o_2, \dots],\ U = [u_1, u_2, \dots]\}$, a set of saved global features and global weights will be randomly drawn from the memory banks $Z_O$ and $Z_U$ with a maximum memory capacity of $t$, concatenated with the current batch, and undergo the decorrelation process together:

$$O = \operatorname{Concat}\big(O, Z_O(i)\big), i = \operatorname{random}(1,2, \dots, t) \tag{15}$$

$$U = \operatorname{Concat}\big(U, Z_U(i)\big) \tag{16}$$

After completing the decorrelation operation for the current batch, the features and weights will be saved back to the memory banks. When the memory banks are not full, the current batch will be directly filled into the memory banks. When the memory banks are full, they are updated as follows:

TABLE 1

DETAILS OF THE DATASETS

| Item | Training | | Testing | |
|---|---|---|---|---|
| | Negative | Positive | Negative | Positive |
| WSIs in Camelyon16 | 154 | 111 | 80 | 49 |
| Patches in Camelyon16 | 1437600 | 1036187 | 737829 | 451919 |
| WSIs in TCGA-NSCLC | 427 | 409 | 107 | 103 |
| Patches in TCGA-NSCLC | 1683444 | 1524389 | 440289 | 423794 |

$$Z'_O(i) = \alpha_i * O[1{:}k] + (1 - \alpha_i) Z_O(i) \tag{17}$$

$$Z'_U(i) = \alpha_i * U[1{:}k] + (1 - \alpha_i) Z_U(i) \tag{18}$$

$$\alpha_i = i/t \tag{19}$$

Here, we use $t$ different smoothing parameters $(\alpha_1, \alpha_2, \dots, \alpha_t)$ to represent the long and short term memory of the memory banks. Subsequently, $(Z_O(i), Z_U(i))$ will be updated by $(Z'_O(i), Z'_U(i))$. During the testing phase, we assume that the features and weights in the memory banks are very similar to the distribution in the real world, so we only draw samples from the memory banks and no longer update them.

### 3.4 Feature Fusion and WSI Classification

The reweighted features $\{o'_1, o'_2, \dots, o'_k\}$ will be fused into WSI-level features and used to construct a WSI classifier. Notably, our CI-MIL is not restricted to any particular feature aggregator, including architecture and training paradigm. The features are fused and used for classification according to the following formula:

$$\hat{Y}_{final} = MLP\left[f_{fusion}(o'_1, o'_2, \dots, o'_k)\right] \tag{20}$$

$$loss_3 = Y * log\left(\hat{Y}_{final}\right) + (1 - Y) * log\left(1 - \hat{Y}_{final}\right) \tag{21}$$

where $\hat{Y}_{final}$ is the final prediction result of CI-MIL and "MLP" represents a multilayer perceptron.

## 4. RESULTS

In this chapter, we first describe the datasets, evaluation metrics, and baselines. Then, we report the classification performance of CI-MIL and the baselines. Next, we report ablation experiments to demonstrate the effectiveness of the two modules dealing with causal relationships proposed in this study. For these two main modules, we evaluated them from both key parameters and model effectiveness. Finally, we analyze the applicability of CI-MIL under different external conditions. Through comprehensive analysis, we demonstrate that CI-MIL improves the accuracy and interpretability of classification by optimizing the construction of WSI prediction results and the real causal relationship with ground truth in the instance-level feature space.

### 4.1 Datasets and Data Preprocessing

Two publicly available datasets representing different diseases were used for model training and evaluation: Camelyon16 and TCGA-NSCLC. Camelyon16 is a WSI dataset used for the detection of breast cancer metastasis. The dataset consists of 400 WSIs (one of which was excluded due to low image quality). WSIs with light background removed [2] were divided into patches of size 256×256-pixel at a 20× magnification. In total, there are about 2.8 million patches. TCGA-NSCLC includes two subtypes of non-small cell lung cancer: lung adenocarcinoma and lung squamous cell carcinoma. This dataset includes 1054 WSIs, which are further divided into about 5.2 million patches at a 20× magnification. These patches are encoded into feature vectors using ResNet18 [40] pre-trained on ImageNet [41], and ResNet18 does not participate in the subsequent model training. The dataset is divided into training and test sets at the WSI level. As shown in Table 1, 129 WSIs in Camelyon16 are divided into the test set, and 210 WSIs in TCGA-NSCLC are divided into the test set.

It is worth noting that the division and preprocessing of the dataset are consistent with the mainstream research in this field, facilitating comparison. For the division of Camelyon16, we referred to the official guidelines. For the division of TCGA-NSCLC, we referred to [19-23]. Using the same tissue extraction algorithm and pre-trained encoder can provide the same external environment for research in this field, reducing computational costs while emphasizing the advantages of the MIL architectures.

TABLE 2

PERFORMANCE COMPARISONS ON THE WSI CLASSIFICATION TASKS

| Mthods | Camelyon16 | | | | TCGA-NSCLC | | | |
|---|---|---|---|---|---|---|---|---|
| | ACC | AUC | Recall | Precision | ACC | AUC | Recall | Precision |
| MaxPooling-MIL (2019) | 81.40 | 79.29 | 65.31 | 82.05 | 81.43 | 81.09 | 63.11 | 98.48 |
| RNN-MIL (2019) | 82.95 | 83.88 | 63.27 | 88.57 | 82.38 | 82.06 | 65.05 | **98.53** |
| CLAM-SB (2021) | 86.05 | 86.82 | 73.47 | 87.80 | 89.52 | 89.36 | 80.58 | 97.65 |
| CLAM-MB (2021) | 86.82 | 85.10 | 69.39 | 94.44 | 85.71 | 88.42 | 72.82 | 97.40 |
| ABMIL (2018) | 84.50 | 84.07 | 81.63 | 78.43 | 81.43 | 88.95 | 85.47 | 78.57 |
| TransMIL (2021) | 83.72 | 81.29 | 81.63 | 76.92 | 85.24 | 90.70 | 85.47 | 84.62 |
| DTFD-MaxS (2022) | 82.95 | 82.77 | 80.09 | 76.74 | 81.90 | 88.91 | 83.77 | 80.37 |
| Ca-ABMIL (2024) | 86.05 | 83.21 | 71.43 | 89.74 | 89.05 | 88.93 | 82.53 | 94.44 |
| Ca-TransMIL (2024) | 89.15 | 88.09 | 83.67 | 89.13 | 90.48 | 90.35 | 83.50 | 96.63 |
| IB-DTFD-MaxS (2023) | 88.37 | 89.51 | 86.51 | 84.00 | 82.86 | 90.5 | 82.96 | 82.52 |
| IB-ABMIL (2023) | 88.37 | 90.43 | 87.14 | 82.69 | 85.24 | 91.26 | 85.44 | 84.62 |
| IB-TransMIL (2023) | 83.72 | 88.71 | 82.93 | 75.93 | 85.24 | 92.54 | 87.06 | 83.33 |
| CI-MIL-AB (Ours) | 90.70 ↑6.20 | **95.41** ↑11.34 | **89.80** ↑8.17 | 86.27 ↑7.84 | 93.33 ↑11.9 | 97.68 ↑8.73 | 94.17 ↑8.7 | 92.38 ↑13.81 |
| CI-MIL-Trans (Ours) | **92.25** ↑8.53 | 95.28 ↑13.99 | 81.63 0.00 | **97.56** ↑20.64 | **94.29** ↑9.05 | **98.07** ↑7.37 | **95.15** ↑9.68 | 93.33 ↑8.71 |

Specifically, the ResNet18 pre-trained on ImageNet is widely used and is considered a low-performance image encoder, making the external environment demanding and non-stationary. Models trained in this way are more reliable for clinical use with many interference factors. Improving the performance of the image encoder by altering the architecture or training paradigm can also enhance the accuracy of subsequent networks in WSI classification, especially when using the corresponding pathological image pre-trained encoder. To rigorously analyze the advantages of CI-MIL, we also referred to [42], compared the performance of different WSI classifiers under the conditions of using pathological images and training encoders, please refer to Section 4.9.

### 4.2 Baselines and Evaluation Metrics

Twelve MIL-based algorithms were selected for comparison. We referenced two feature fusion MIL methods, ABMIL (Attention Based MIL) [19] and TransMIL (Transformer MIL) [20], to construct CI-MIL-AB and CI-MIL-Trans. The reason for choosing these two methods as the basis is that they are two classic feature aggregators, and [18, 21-23] are all developed based on them. The other 10 algorithms are MaxPooling-MIL [1], RNN-MIL [1], CLAM- SB [2], CLAM-MB [2], DTFD-MaxS [18], CaMIL-ABMIL [22], CaMIL-TransMIL [22], IBMIL-DTFD [23], IBMIL-ABMIL [23], and IBMIL-TransMIL [23]. All baselines are configured according to official settings, if available. In addition, our replication results for the baselines are the same or similar to those in [20-23], which is convincing.

For the WSI classification task, accuracy (ACC) and the area under the receiver operating characteristic curve (AUC) are the most important evaluation metrics. To analyze these methods more comprehensively, recall and precision were also selected as evaluation metrics.

### 4.3 WSI Classification Results

Comparing with multiple baselines under the same experimental conditions and displaying metrics is the most intuitive way to evaluate new approaches. Table 2 shows the classification results of the algorithms on Camelyon16 and TCGA-NSCLC, respectively. The green upward arrow represents an improvement in performance over the original method, while grey represents no performance improvement. In the classification task of Camelyon16, CI-MILs have ACC and AUC that surpasses all existing methods. At the same time, the performance improvement of CI-MIL over the original methods is considerable. CI-MIL improves the accuracy of ABMIL by 6.2%, AUC by 11.34%, and recall by 8.17%. In terms of improving TransMIL, the improvement in AUC reached an astonishing 13.99%. At the same time, ACC and precision have seen considerable increases of 8.53% and 12.13%, respectively. On TCGA-NSCLC, CI-MILs achieved a maximum of 94.29% ACC and 98.07% AUC, surpassing the baselines with

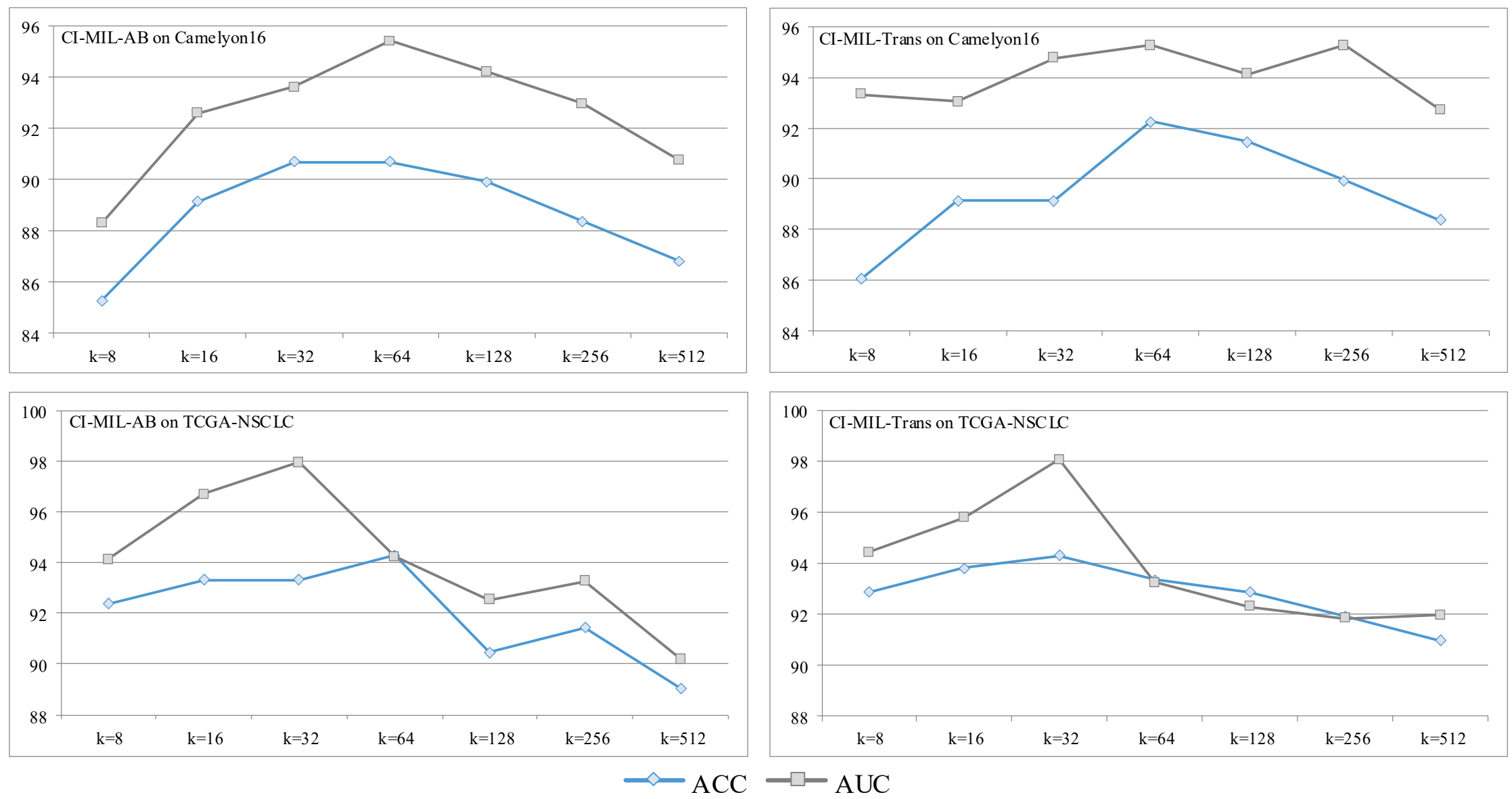

Fig. 4. The impact of feature distillation scale on classification performance.

TABLE 3

ABLATION EXPERIMENTS

| Components | | CI-MIL-AB | | | | CI-MIL-Trans | | | |
|---|---|---|---|---|---|---|---|---|---|
| | | Camelyon16 | | TCGA-NSCLC | | Camelyon16 | | TCGA-NSCLC | |
| Stage 1 | Stage 2 | ACC | AUC | ACC | AUC | ACC | AUC | ACC | AUC |
| × | × | 84.5 | 84.07 | 81.43 | 88.95 | 83.72 | 81.29 | 85.24 | 90.7 |
| √ | × | 86.05 | 89.01 | 84.29 | 83.4 | 85.27 | 93.41 | 88.1 | 92.38 |
| × | √ | 87.6 | 92.72 | 90.95 | 91.87 | 87.6 | 94.11 | 91.43 | 93.35 |
| √ | √ | 90.70 | 95.41 | 93.33 | 97.98 | 92.25 | 95.28 | 94.29 | 98.07 |

a considerable advantage. At the same time, CI-MIL also produced significant performance improvements over the original methods for all evaluation metrics.

Compared with the two types of methods that build causal relationships through feature clustering: CaMILs and IB-MILs, CI-MIL is not only a universal model orthogonal to mainstream feature aggregators, but also far exceeds CaMILs and IB-MILs in performance. This indicates that our designed patch-feature decorrelation method, compared to traditional feature clustering methods, can construct a causal relationship closer to the real-world distribution, thus having better classification performance.

### 4.4 Ablation Experiments

We verify the effectiveness of the two-stage method of processing correlation in CI-MIL through ablation experiments: the first stage of feature distillation and the second stage of feature decorrelation. Evaluations were conducted under four conditions: 1. No decorrelation in both stages (i.e., traditional ABMIL and TransMIL); 2. Only the first stage; 3. Only the second stage; 4. Complete CI-MILs. The ACC and AUC under each condition are reported, as shown in Table 3.

The results show that both stages of processing correlation operations improve classification performance (Table 3). At the same time, the effect of feature decorrelation operation in random Fourier space is superior. The average improvement for ACC is 5.67%, and for AUC is 6.76%, exceeding the corresponding 2.21% and 3.3% of feature distillation. This result proves that the patch-level feature decorrelation operation in random Fourier space designed in this study plays a greater role in reducing false correlations and establishing real dependencies between predictions and ground truth. Therefore, it shows better performance compared to traditional causal inference studies based on feature clustering. Through this comparison, both stages have been proven effective in constructing real dependencies between predictions and evidences, enhancing classification performance.

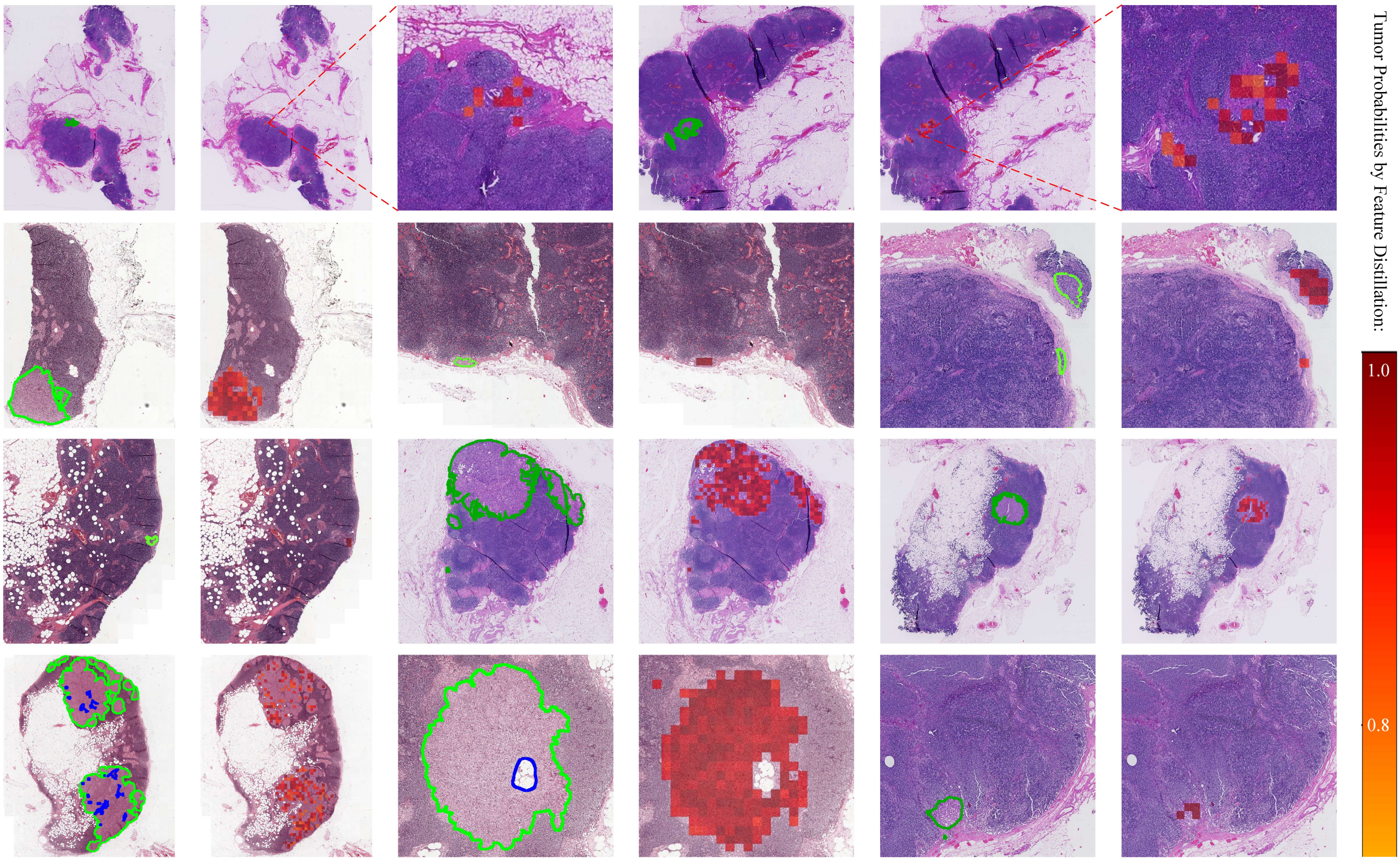

Fig. 5. Visually compare ground truth and the distilled diagnostic correlation sub-regions. The area enclosed by the green curve but not by the blue curve is the ground truth, and the red patches represent the area distilled by CI-MIL.

### 4.5 The Impact of Feature Distillation Scale on Classification Performance

The purpose of feature distillation proposed in this study is to trim the flat part of the patch-level feature hyperplane, thereby initially establishing a connection between model predictions and diagnostic evidence sub-region images. The distillation scale is a crucial hyperparameter. Each WSI contains thousands of patches. If the scale is too small, it cannot reflect sufficient diagnostic information. Conversely, if the scale is too large, it will bring computational burden and introduce additional interference. The distillation scale has been represented in Section 3.2 by the number of patch-level features distilled from each WSI, denoted as "$k$". We evaluated the change in classification performance with the change in the $k$ value, as shown in Fig.4. The adjustment of hyperparameters is applied to both the training sets and testing sets.

The results indicate that when the distillation scale is in the order of tens of patches, the model exhibits higher performance. The two datasets involved in this study have different image areas. For Camelyon16, each WSI contains an average of about 10,000 patches, while TCGA-NSCLC contains about 4,000. The optimal distillation scale on these two datasets also differs, with the optimal k value chosen for Camelyon16 being 64, slightly higher than the 32 for TCGA-NSCLC. This also indicates that diagnostic evidence patches usually only account for a small portion of the WSI, or can be described as a certain proportion of the image area. Preliminary positioning of these patches can establish a correlation between model predictions and evidences. There is no need to include more patches, which are redundant and contain interference for model predictions. The feature distillation introduced in this study effectively solves this problem.

### 4.6 Visually Compare Ground Truth and The Distilled Diagnostic Correlation Sub-regions

To intuitively evaluate the ability of the feature distillation to construct true correlation between model predictions and diagnostic evidence patches, we plotted heatmaps of the regions of interest (ROIs) in feature distillation to visually compare with the manual pixelwise annotations by pathologists.

The Camelyon16 dataset contains pixelwise annotations for all positive WSIs, so we selected part of its test set for evaluation. The results are shown in Fig.5, where the green curve outlines the tumor cells as identified by the pathologist; the red patches

TABLE 4

DIFFERENT METHODS OF CORRELATION MATRIX CONSTRUCTION

| Methods | | Camelyon16 | | TCGA-NSCLC | |
|---|---|---|---|---|---|
| | | ACC | AUC | ACC | AUC |
| CI-MIL-AB | $[Cov]$ | 86.1 | 95.2 | 89.1 | 92.9 |
| | $[In]$ | 89.9 | 95.1 | 91.0 | 92.8 |
| | $[Cov]+[In]$ | 90.7 | 95.4 | 93.3 | 98.0 |
| CI-MIL-Trans | $[Cov]$ | 89.2 | 94.2 | 92.9 | 96.8 |
| | $[In]$ | 88.4 | 94.1 | 91.9 | 95.9 |
| | $[Cov]+[In]$ | 92.3 | 95.3 | 94.3 | 98.1 |

represent the ROIs of feature distillation, and the depth of the red color indicates the tumor probability as perceived by the model. It can be clearly observed that there is a significant overlap between the areas distilled by CI-MIL and the ground truth. The areas selected by CI-MIL are almost entirely encompassed by the curves drawn by the pathologist, demonstrating that the correlation between prediction and evidence established by feature distillation is similar to human thinking. The occurrence of false correlations where "the prediction results are associated with non-tumor images" is reduced. Also, due to the distillation scale discussed in Section 4.5, CI-MIL's ROIs may not fully cover the tumor area, but the distilled patches are already sufficient to establish real correlations and complete predictions. CI-MIL demonstrates its ability to accurately capture tumor areas in WSIs, thereby improving interpretability in classification tasks. This result proves that this method has more potential applications in tasks such as tumor detection and diagnostic assistance, not limited to classification. For example, CI-MIL can provide diagnostic clues to pathologists through localization, improving human diagnostic efficiency and accuracy.

### 4.7 Evaluate Different Methods of Correlation Matrix Construction

We explored the impact of different correlation matrix construction methods discussed in Section 3.3. Two correlation matrix construction methods are discussed, namely $[Cov]$ based on covariance and $[In-product]$ based on the angle between spatial vectors. Particularly for $[In-product]$, it reduces the inter-correlation of patch-level features by forcing all feature vectors in high-dimensional space to be orthogonal to each other. This method is calculated in the form of matrix inner product, which obviously has a smaller computational load compared to calculating the covariance between features. The results, as shown in Table 4, reveal that each of the two correlation matrix calculation methods has its own strengths in terms of performance. When both methods are employed simultaneously, the performance sees a slight improvement. This suggests that the two patch-level feature decorrelation methods introduced in this study can complement each other, better decorrelate patch-level features, and construct a smoother feature hyperplane. This reduces the problem of large feature differences brought about by strong correlation of patch-level features, yet having similar conditional probabilities (i.e., the probability of belonging to positive instances under this feature distribution).

### 4.8 Evaluate The Correlation Between Patch-level Features

We verified that the operation of decorrelation of patch-level features in random Fourier feature space can meet the assumption: by decorrelating all features, a "waveless" patch-level feature hyperplane is constructed. The principle is to eliminate the situation where the feature distribution differences are large due to high feature correlation, but the predicted probabilities are similar. We compared the inter-correlation of patch-level features before and after the decorrelation operation in the training set and test set of the two datasets, a total of four groups of data. For each WSI, we randomly extracted a batch of features. This is to maintain consistency with the parameters during the training process and to examine the global features of the WSI, rather than a small number of specific features processed by feature distillation. Then, the correlation matrices before and after the decorrelation operation within each batch were constructed. The sum of the off-diagonal elements of the matrix was calculated to evaluate the inter-correlation of these features. Finally, the average result of all WSIs in each group of data was calculated. The results are shown in Fig.6. This result indicates that CI-MIL can consistently reduce the inter-correlation of patch features in both the training set and

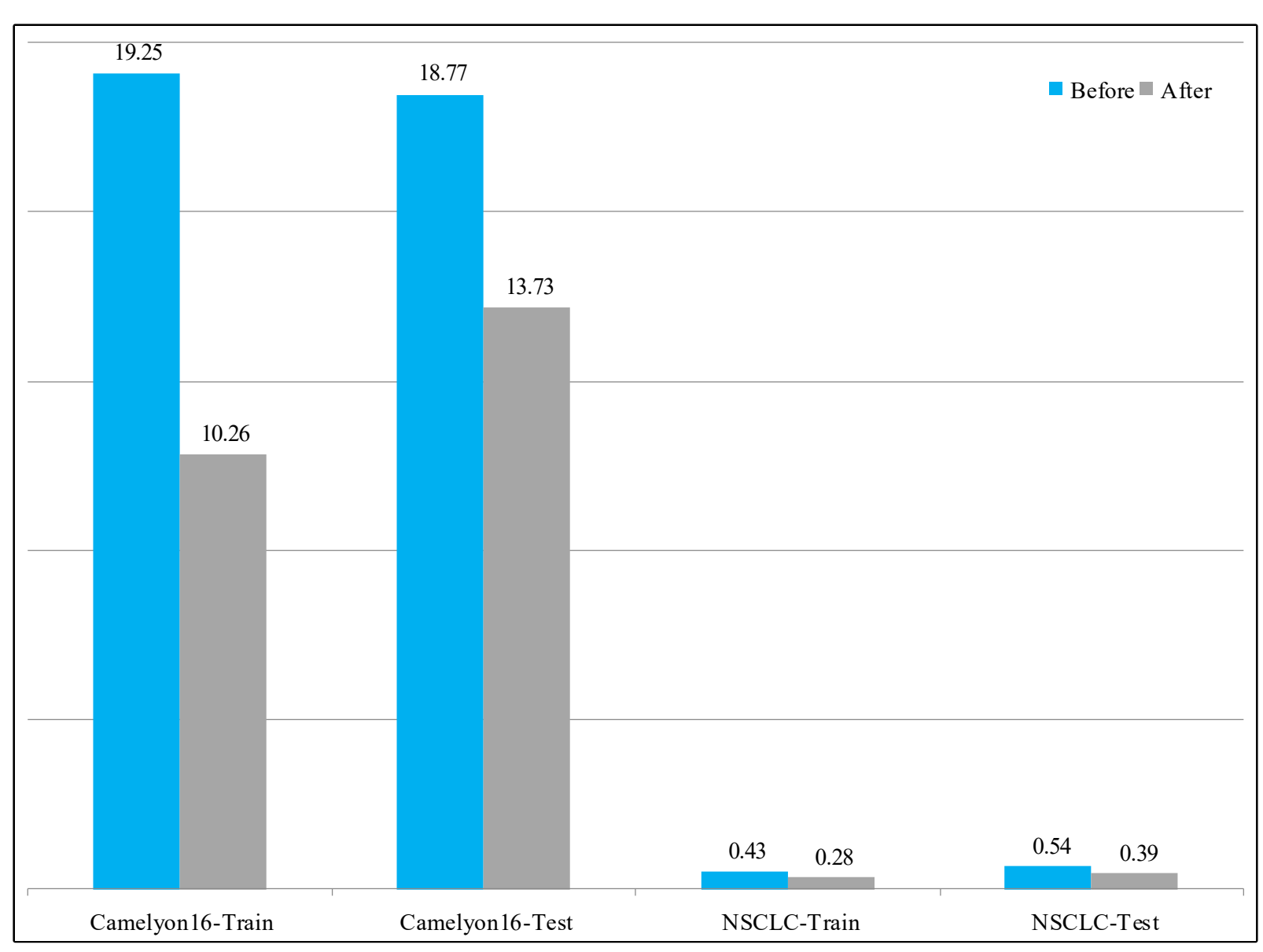

Fig. 6. The correlation between patch-level features is reduced.

TABLE 5

CLASSIFICATION RESULTS COMBINED WITH GCMAE

| Mthods | ACC | AUC | Recall | Precision |
|---|---|---|---|---|
| ABMIL | 88.37 | 91.09 | 71.43 | 97.22 |
| TransMIL | 87.6 | 89.65 | 77.55 | 88.37 |
| IB-ABMIL | 90.7 | 92.94 | 81.63 | 93.02 |
| IB-TransMIL | 91.47 | 94.44 | 79.59 | 97.5 |
| CI-MIL-AB | 96.12 | 97.22 | **95.92** | 96.48 |
| CI-MIL-Trans | **96.12** | **97.46** | 97.96 | 92.31 |

test set, making the feature hyperplane smooth and "waveless". Based on this hyperplane, simple partitioning curves can achieve a similarity to the real world, thereby reducing overfitting and false diagnostic correlations.

### 4.9 Classification Performance Under More Stable External Conditions

As mentioned in Section 4.1, ResNet18 is not a particularly outstanding patch image encoder. Many studies have shown that better encoders will bring higher performance, but this is not thanks to the merit of the MIL framework [22, 23]. Essentially, this is also a kind of optimization of the patch-level feature hyperplane, making it closer to the real-world distribution. We further explored whether CI-MIL can establish a more realistic correlation between predictions and evidence under more stable external conditions brought by encoders, thereby exhibiting safer performance and scalability at the clinical application level. Pre-training the encoder with pathological images is an effective way to make the model learn high-level image information. We referred to GCMAE [42], an encoder for a vision transformer [43] architecture specifically for pathological images trained with a random mask self-supervised strategy. We used the weights applicable to Camelyon16 announced by the official. The performance of CI-MILs and some baselines on Camelyon16 is shown in Table 5. Under this condition, CI-MILs still surpass the baselines, and the performance is further improved. CI-MIL-Trans achieved the highest performance, with an ACC of 96.12% and an AUC of 97.46%. At the same time, CI-MIL-Trans achieved a recall rate of 97.96% for malignant samples, which has important safety implications in clinical applications. In addition, compared to the results in Table 2, when CI-MIL is used as a method to improve the source models, it achieves better performance than improving the encoding environment. This also proves that CI-MIL is a better scheme to optimize the feature hyperplane and is closer to the real-world distribution. Its principle of establishing real causal relationships compares favorably with the mainstream methods at this stage: improving feature aggregators or feature encoders, and can better complete pathological clinical prediction tasks. CI-MIL has been proven to have better applicability whether the external conditions are stable or not.

## 5. CONCLUSION

In this study, we have introduced a novel approach termed Causal Inference Multiple Instance Learning to address the inherent challenges in weakly supervised WSI classification. By incorporating feature distillation and causal inference, CI-MIL effectively reduces the spurious correlation between model predictions and non-diagnostic patches, thus enhancing the stability and interpretability of WSI predictions.

We inducted the concept of patch-level feature hyperplane in WSI classification tasks. Guided by this concept, CI-MIL optimizes the feature hyperplane in a two-stage manner, thereby establishing a real causal relationship between model predictions and diagnostic evidence, as evidenced by the outstanding performance metrics we achieved on the Camelon16 and TCGA-NSCLC datasets. The consistency between the patches selected by CI-MIL and the ground truth annotations further emphasizes the interpretability and reliability of this method. Overall, CI-MIL provides a powerful framework for improving the accuracy and reliability of WSI-based diagnoses. Its universal design allows for seamless integration with existing models, providing valuable tools for enhancing the clinical diagnostic process.

## REFERENCES

[1] Nan, T., Zheng, S., Qiao, S. et al. Deep learning quantifies pathologists' visual patterns for whole slide image diagnosis. Nat Commun 16, 5493 (2025). https://doi.org/10.1038/s41467-025-60307-1

[2] M. Y. Lu, D. F. Williamson, T. Y. Chen, R. J. Chen, M. Barbieri, and F. Mahmood, "Data-efficient and weakly supervised computational pathology on whole-slide images," Nature biomedical engineering, vol. 5, no. 6, pp. 555-570, 2021.

[3] S. I. Hajdu, "Microscopic contributions of pioneer pathologists," Annals of Clinical & Laboratory Science, vol. 41, no. 2, pp. 201-206, 2011.

[4] K. Bera, K. A. Schalper, D. L. Rimm, V. Velcheti, and A. Madabhushi, "Artificial intelligence in digital pathology—new tools for diagnosis and precision oncology," Nature reviews Clinical oncology, vol. 16, no. 11, pp. 703-715, 2019.

[5] M. K. K. Niazi, A. V. Parwani, and M. N. Gurcan, "Digital pathology and artificial intelligence," The lancet oncology, vol. 20, no. 5, pp. e253-e261, 2019.

[6] T. J. Fuchs, P. J. Wild, H. Moch, and J. M. Buhmann, "Computational pathology analysis of tissue microarrays predicts survival of renal clear cell carcinoma patients," in Medical Image Computing and Computer-Assisted Intervention–MICCAI 2008: 11th International Conference, New York, NY, USA, September 6-10, 2008, Proceedings, Part II 11, 2008: Springer, pp. 1-8.

[7] T. J. Fuchs and J. M. Buhmann, "Computational pathology: challenges and promises for tissue analysis," Computerized Medical Imaging and Graphics, vol. 35, no. 7-8, pp. 515-530, 2011.

[8] D. N. Louis et al., "Computational pathology: a path ahead," Archives of pathology & laboratory medicine, vol. 140, no. 1, pp. 41-50, 2016.

[9] D. Schapiro et al., "histoCAT: analysis of cell phenotypes and interactions in multiplex image cytometry data," Nature methods, vol. 14, no. 9, pp. 873-876, 2017.

[10] E. Moen, D. Bannon, T. Kudo, W. Graf, M. Covert, and D. Van Valen, "Deep learning for cellular image analysis," Nature methods, vol. 16, no. 12, pp. 1233-1246, 2019.

[11] R. Pell et al., "The use of digital pathology and image analysis in clinical trials," The Journal of Pathology: Clinical Research, vol. 5, no. 2, pp. 81-90, 2019.

[12] Z.-H. Zhou, "A brief introduction to weakly supervised learning," National science review, vol. 5, no. 1, pp. 44-53, 2018.

[13] O. Maron and T. Lozano-Pérez, "A framework for multiple-instance learning," Advances in neural information processing systems, vol. 10, 1997.

[14] Y. LeCun, Y. Bengio, and G. Hinton, "Deep learning," nature, vol. 521, no. 7553, pp. 436-444, 2015.

[15] A. Esteva et al., "Dermatologist-level classification of skin cancer with deep neural networks," nature, vol. 542, no. 7639, pp. 115-118, 2017.

[16] R. Poplin et al., "Prediction of cardiovascular risk factors from retinal fundus photographs via deep learning," Nature biomedical engineering, vol. 2, no. 3, pp. 158-164, 2018.

[17] S. M. McKinney et al., "International evaluation of an AI system for breast cancer screening," Nature, vol. 577, no. 7788, pp. 89-94, 2020.

[18] B. Li, Y. Li, and K. W. Eliceiri, "Dual-stream multiple instance learning network for whole slide image classification with self-supervised contrastive learning," in Proceedings of the IEEE/CVF conference on computer vision and pattern recognition, 2021, pp. 14318-14328.

[19] M. Ilse, J. Tomczak, and M. Welling, "Attention-based deep multiple instance learning," in International conference on machine learning, 2018: PMLR, pp. 2127-2136.

[20] Z. Shao, H. Bian, Y. Chen, Y. Wang, J. Zhang, and X. Ji, "Transmil: Transformer based correlated multiple instance learning for whole slide image classification," Advances in neural information processing systems, vol. 34, pp. 2136-2147, 2021.

[21] H. Zhang et al., "Dtfd-mil: Double-tier feature distillation multiple instance learning for histopathology whole slide image classification," in Proceedings of the IEEE/CVF conference on computer vision and pattern recognition, 2022, pp. 18802-18812.

[22] K. Chen, S. Sun, and J. Zhao, "Camil: Causal multiple instance learning for whole slide image classification," in Proceedings of the AAAI Conference on Artificial Intelligence, 2024, vol. 38, no. 2, pp. 1120-1128.

[23] T. Lin, Z. Yu, H. Hu, Y. Xu, and C.-W. Chen, "Interventional bag multi-instance learning on whole-slide pathological images," in Proceedings of the IEEE/CVF Conference on Computer Vision and Pattern Recognition, 2023, pp. 19830-19839.

[24] X. Yang, H. Zhang, G. Qi, and J. Cai, "Causal attention for vision-language tasks," in Proceedings of the IEEE/CVF conference on computer vision and pattern recognition, 2021, pp. 9847-9857.

[25] K. Kuang, R. Xiong, P. Cui, S. Athey, and B. Li, "Stable prediction with model misspecification and agnostic distribution shift," in Proceedings of the AAAI Conference on Artificial Intelligence, 2020, vol. 34, no. 04, pp. 4485-4492.

[26] Z. Shen, P. Cui, T. Zhang, and K. Kunag, "Stable learning via sample reweighting," in Proceedings of the AAAI Conference on Artificial Intelligence, 2020, vol. 34, no. 04, pp. 5692-5699.

[27] K. Kuang, P. Cui, S. Athey, R. Xiong, and B. Li, "Stable prediction across unknown environments," in proceedings of the 24th ACM SIGKDD international conference on knowledge discovery & data mining, 2018, pp. 1617-1626.

[28] B. E. Bejnordi et al., "Diagnostic assessment of deep learning algorithms for detection of lymph node metastases in women with breast cancer," Jama, vol. 318, no. 22, pp. 2199-2210, 2017.

[29] K. Zarogoulidis et al., "Treatment of non-small cell lung cancer (NSCLC)," Journal of thoracic disease, vol. 5, no. Suppl 4, p. S389, 2013.

[30] P.-H. C. Chen et al., "An augmented reality microscope with real-time artificial intelligence integration for cancer diagnosis," Nature medicine, vol. 25, no. 9, pp. 1453-1457, 2019.

[31] K. Das, S. P. K. Karri, A. G. Roy, J. Chatterjee, and D. Sheet, "Classifying histopathology whole-slides using fusion of decisions from deep convolutional network on a collection of random multi-views at multi-magnification," in 2017 IEEE 14th International Symposium on Biomedical Imaging (ISBI 2017), 2017: IEEE, pp. 1024-1027.

[32] M. Valkonen, K. Kartasalo, K. Liimatainen, M. Nykter, L. Latonen, and P. Ruusuvuori, "Metastasis detection from whole slide images using local features and random forests," Cytometry Part A, vol. 91, no. 6, pp. 555-565, 2017.

[33] B. Ehteshami Bejnordi et al., "Using deep convolutional neural networks to identify and classify tumor-associated stroma in diagnostic breast biopsies," Modern Pathology, vol. 31, no. 10, pp. 1502-1512, 2018.

[34] P. Mobadersany et al., "Predicting cancer outcomes from histology and genomics using convolutional networks," Proceedings of the National Academy of Sciences, vol. 115, no. 13, pp. E2970-E2979, 2018.

[35] A. Janowczyk and A. Madabhushi, "Deep learning for digital pathology image analysis: A comprehensive tutorial with selected use cases," Journal of pathology informatics, vol. 7, no. 1, p. 29, 2016.

[36] G. Litjens et al., "Deep learning as a tool for increased accuracy and efficiency of histopathological diagnosis," Scientific reports, vol. 6, no. 1, p. 26286, 2016.

[37] D. Zhang, H. Zhang, J. Tang, X.-S. Hua, and Q. Sun, "Causal intervention for weakly-supervised semantic segmentation," Advances in Neural Information Processing Systems, vol. 33, pp. 655-666, 2020.

[38] X. Zhang, P. Cui, R. Xu, L. Zhou, Y. He, and Z. Shen, "Deep stable learning for out-of-distribution generalization," in Proceedings of the IEEE/CVF Conference on Computer Vision and Pattern Recognition, 2021, pp. 5372-5382.

[39] Geirhos, R., Jacobsen, J. H., Michaelis, C., Zemel, R., Brendel, W., Bethge, M., & Wichmann, F. A. (2020). Shortcut learning in deep neural networks. Nature Machine Intelligence, 2(11), 665-673.

[40] K. He, X. Zhang, S. Ren, and J. Sun, "Deep residual learning for image recognition," in Proceedings of the IEEE conference on computer vision and pattern recognition, 2016, pp. 770-778.

[41] J. Deng, W. Dong, R. Socher, L.-J. Li, K. Li, and L. Fei-Fei, "Imagenet: A large-scale hierarchical image database," in 2009 IEEE conference on computer vision and pattern recognition, 2009: Ieee, pp. 248-255.

[42] H. Quan et al., "Global Contrast-Masked Autoencoders Are Powerful Pathological Representation Learners," Pattern Recognition, p. 110745, 2024.

[43] K. Han et al., "A survey on vision transformer," IEEE transactions on pattern analysis and machine intelligence, vol. 45, no. 1, pp. 87-110, 2022.